# Using Text Analytics for Health to Get Meaningful Insights from a Corpus of COVID Scientific Papers

Dmitry Soshnikov[1,2], Vickie Soshnikova[3]

[1] Microsoft Developer Relations
[2] Higher School of Economics, Moscow, Russia
[3] Phystech Lyceum, Dolgoprudny, Russia
Correspondence address: dmitri@soshnikov.com

**Abstract:** Since the beginning of COVID pandemic, there have been around 700000 scientific papers published on the subject. A human researcher cannot possibly get acquainted with such a huge text corpus—and therefore developing AI-based tools to help navigating this corpus and deriving some useful insights from it is highly needed. In this paper, we will use Text Analytics for Health pre-trained service together with some cloud tools to extract some knowledge from scientific papers, gain insights, and build a tool to help researcher navigate the paper collection in a meaningful way.

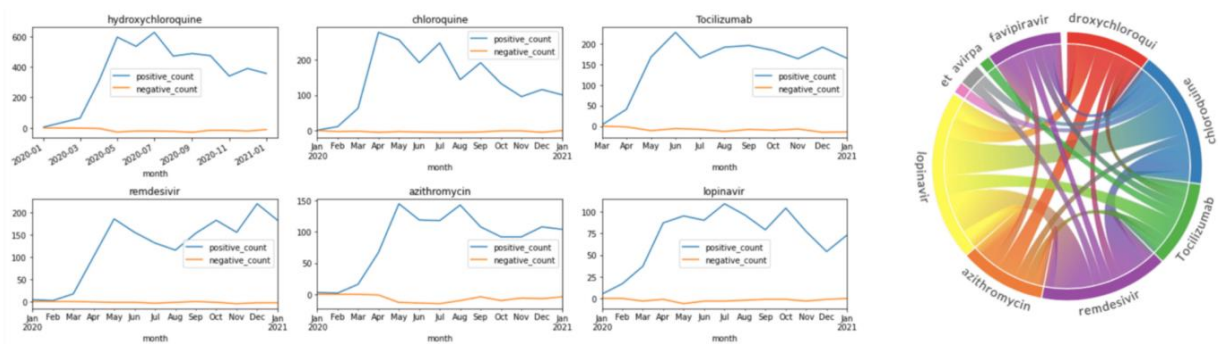

**Code.** The code for experiments described in this paper is available from http://github.com/CloudAdvocacyAzurePaperExplorationWorkshop.

## Introduction: Automatic Paper Analysis

Automatic scientific paper analysis is fast growing area of studies, and due to recent improvements in NLP techniques is has been greatly improved in the recent years. In this paper, we will show you how to derive specific insights from COVID papers, such as changes in medical treatment over time, or joint treatment strategies using several medications.

The main approach we will describe in this paper is to extract as much semi-structured information from text as possible, and then store it into some NoSQL database for further processing. Storing information in the database would allow us to make some very specific queries to answer some of the questions, as well as to provide visual exploration tool for medical expert for structured search and insight generation. Processing large volume of papers is done by running parallel sweep job on Azure Machie Learning cluster.

The overall architecture of the proposed system is shown in Fig. 1:

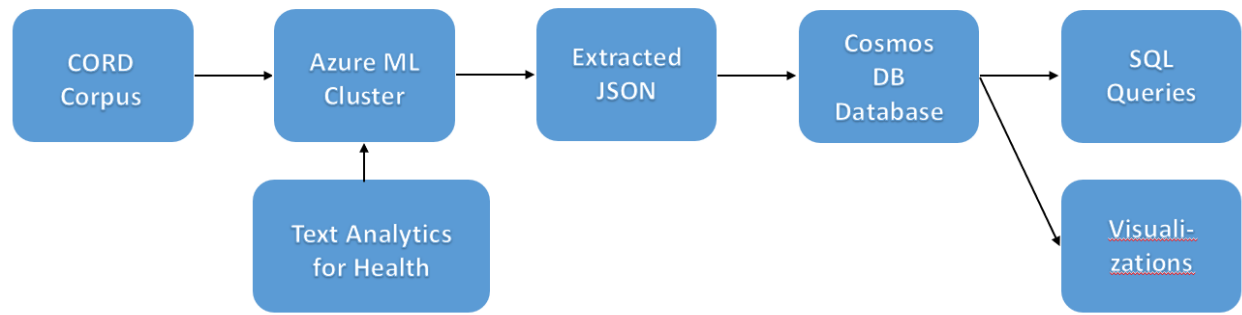

**Fig.1:** Architecture of a system to extract knowledge insights from a corpus of scientific papers. Note that this architecture is built on top of the platform components of Microsoft Azure, which allows us to delegate many complex issues (such as scalability) to the cloud provider.

## COVID Scientific Papers and CORD Dataset

The idea to apply NLP methods to scientific literature seems quite natural and has been proposed in many different works [1,2,3]. First of all, scientific texts are already well-structured, they contain things like keywords, abstract, as well as well-defined terms. Thus, at the very beginning of COVID pandemic, a research challenge has been launched on Kaggle to analyze scientific papers on the subject. The dataset behind this competition is called CORD [4], and it contains constantly updated corpus of everything that is published on topics related to COVID.

This dataset consists of the following parts:

- **Metadata file [Metadata.csv](Metadata.csv)** contains most important information for all publications in one place. Each paper in this table has unique identifier `cord_uid` (which in fact does not happen to be completely unique, once you start working with the dataset). The information includes: Title of publication, Journal, Authors, Abstract, Date of publication, doi
- **Full-text papers** in `document_parses` directory that contain structured text in JSON format, which greatly simplifies the analysis.
- Pre-built **Document Embeddings** that maps `cord_uid`s to float vectors that reflect some overall semantics of the paper.

In this paper, we will focus on paper abstracts, because they contain the most important information from the paper. However, for full analysis of the dataset, it makes sense to use the same approach on full texts as well.

## Natural Language Processing Tasks

In the recent years, there has been a huge progress in the field of Natural Language Processing, and very powerful neural network language models have been trained. In the area of NLP, the following tasks are typically considered:

- **Text classification / intent recognition**—In this task, we need to classify a piece of text into a number of categories. This is a typical classification task.
- **Sentiment Analysis**—We need to return a number that shows how positive or negative the text is. This is a typical regression task.
- **Named Entity Recognition** (NER)—In NER, we need to extract named entities from text, and determine their type. For example, we may be looking for names of medicines, or diagnoses. Another task similar to NER is **keyword extraction**.

- **Text summarization**—Here we want to be able to produce a short version of the original text, or to select the most important pieces of text.
- **Question Answering**—In this task, we are given a piece of text and a question, and our goal is to find the exact answer to this question from text.
- **Open-Domain Question Answering** (ODQA)—The main difference from previous task is that we are given a large corpus of text, and we need to find the answer to our question somewhere in the whole corpus.

In [5] have described how we can use ODQA approach to automatically find answers to specific COVID questions. However, this approach does not provide insights into the text corpus.

To make some insights from text, NER seems to be the most prominent technique to use. If we can find specific entities that are present in text, we could then perform semantically rich search in text that answers specific questions, as well as obtain data on co-occurrence of different entities, figuring out specific scenarios that interest us.

To train NER model, as well as any other neural language model, we need a reasonably large dataset that is properly marked up. Finding those datasets is often not an easy task, and producing them for new problem domain often requires initial human effort to mark up the data.

# Pre-Trained Language Models and Text Analytics for Health Cognitive Service

Luckily, modern transformer language models can be trained in semi-supervised manner using transfer learning. First, the base language model (for example, BERT [6]) is trained on a large corpus of text first, and then can be specialized to a specific task such as classification or NER on a smaller dataset.

This transfer learning process can also contain additional step—further training of generic pre-trained model on a domain-specific dataset. For example, in the area of medical science Microsoft Research has pre-trained a model called PubMedBERT [7], using texts from PubMed repository. This model can then be further adopted to different specific tasks, provided we have some specialized datasets available.

However, training a model requires a lot of skills and computational power, in addition to a dataset. Microsoft (as well as some other large cloud vendors) also make some pre-trained models available through the REST API. Those services are called Cognitive Services, and one of those services for working with text is called Text Analytics [8]. It can do the following:

- **Keyword extraction** and NER for some common entity types, such as people, organizations, dates/times, etc.
- **Sentiment analysis**
- **Language Detection**
- **Entity Linking**, by automatically adding internet links to some most common entities. This also performs **disambiguation**, for example *Mars* can refer to both the planet or a chocolate bar, and correct link would be used depending on the context.

For example, here is the result of analyzing one medical paper abstract by Text Analytics:

As a result of the 2019 `DateTime` coronavirus disease pandemic `Event` ( COVID-19) `Event` , there has been an urgent worldwide demand for treatments. Due to factors such as history `Skill` of prescription for other infectious `Skill` diseases, availability, `Skill` and relatively low cost, the use of chloroquine `Product` (CQ) and hydroxychloroquine `Product` (HCQ) has been tested in vivo and in vitro for the ability to inhibit the causative virus, severe acute respiratory syndrome coronavirus 2 `Quantity` (SARS-CoV-2). However, even though investigators noted the therapeutic potential of these drugs, it is important to consider the toxicological risks and necessary care for rational use of CQ and HCQ. This study provides information on the main toxicological and epidemiological aspects to be considered for prophylaxis or treatment of COVID-19 using CQ but mainly HCQ, which is a less toxic derivative than CQ, and was shown to produce better results in inhibiting proliferation of SARS-CoV-2 based upon preliminary tests.

As you can see, some specific entities (for example, HCQ, which is short for hydroxychloroquine) are not recognized at all.

Recently, a special version of the service, called Text Analytics for Health [9] was released, which exposes pre-trained PubMedBERT model with some additional capabilities. Here is the result of extracting entities from the same piece of text using Text Analytics for Health:

As a result of the 2019 `Time` coronavirus disease pandemic `Diagnosis` ( COVID-19) `Diagnosis` , there has been an urgent worldwide demand for treatments. `TreatmentName` Due to factors such as history of prescription for other infectious diseases, `Diagnosis` availability, and relatively low cost, the use of chloroquine `MedicationName` ( CQ) `MedicationName` and hydroxychloroquine `MedicationName` ( HCQ) `MedicationName` has been tested in vivo and in vitro for the ability to inhibit the causative virus, severe acute respiratory syndrome coronavirus 2 `Diagnosis` ( SARS-CoV-2 `Diagnosis` ). However, even though investigators noted the therapeutic potential of these drugs, it is important to consider the toxicological risks and necessary care for rational use of CQ `MedicationName` and HCQ. `MedicationName` This study provides information on the main toxicological and epidemiological aspects to be considered for prophylaxis `TreatmentName` or treatment `TreatmentName` of COVID-19 `Diagnosis` using CQ `MedicationName` but mainly HCQ, `MedicationName` which is a less toxic `SymptomOrSign` derivative than CQ, `MedicationName` and was shown to produce better results in inhibiting proliferation of SARS-CoV-2 `Diagnosis` based upon preliminary tests.

Text Analytics is a REST service, which can be called by using Text Analytics Python SDK in the following manner:

```
poller = text_analytics_client.begin_analyze_healthcare_entities([txt])
res = list(poller.result())
print(res)
```

In addition to just the list of entities, we also get the following:

- **Enity Mapping** of entities to standard medical ontologies, such as UMLS [10].
- **Relations** between entities inside the text, such as `TimeOfCondition`, etc.
- **Negation**, which indicated that an entity was used in negative context, for example *COVID-19 diagnosis did not occur*.

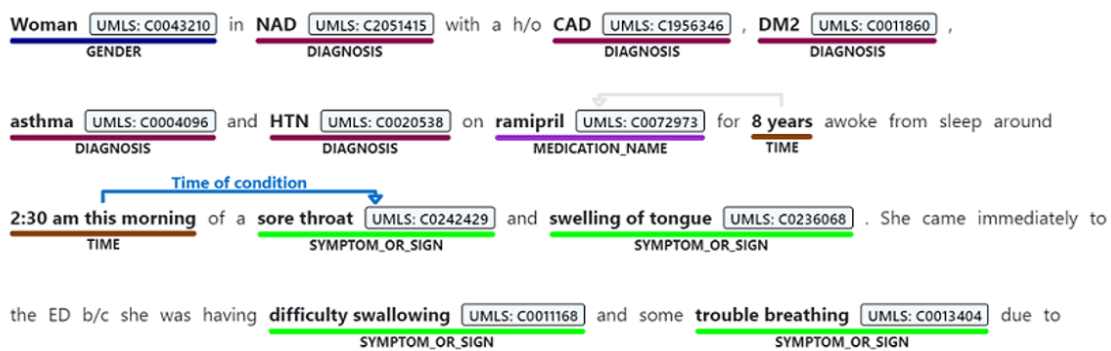

**Fig. 2:** Results of entity extraction, linking and ontology mapping returned by Text Analytics for Health

In addition to using Python SDK, we can also call Text Analytics using REST API directly. This is useful if you are using a programming language that does not have a corresponding SDK, or if you prefer to receive Text Analytics result in the JSON format for further storage or processing. In Python, this can be easily done using `requests` library:

```python
uri = f"{endpoint}/text/analytics/v3.1/entities/
        health/jobs?model-version=v3.1"
headers = { "Ocp-Apim-Subscription-Key" : key }
resp = requests.post(uri,headers=headers,data=doc)
res = resp.json()
if res['status'] == 'succeeded':
    result = t['results']
else:
    result = None
```

Resulting JSON file will look like this:

```
{"id": "jk62qn0z",
 "entities": [
    {"offset": 24, "length": 28, "text": "coronavirus disease pandemic",
     "category": "Diagnosis", "confidenceScore": 0.98,
     "isNegated": false},
    {"offset": 54, "length": 8, "text": "COVID-19",
     "category": "Diagnosis", "confidenceScore": 1.0, "isNegated": false,
     "links": [
        {"dataSource": "UMLS", "id": "C5203670"},
        {"dataSource": "ICD10CM", "id": "U07.1"}, ... ]},
 "relations": [
    {"relationType": "Abbreviation", "bidirectional": true,
     "source": "#/results/documents/2/entities/6",
     "target": "#/results/documents/2/entities/7"}, ...],
}
```

In production code, one may want to incorporate a mechanism that will retry the operation when an error is returned by the service.

## Parallel Paper Processing using Azure Machine Learning Cluster

Since the dataset currently contains ~700K paper abstracts, processing them sequentially through Text Analytics would be quite time-consuming. To run this code in parallel, we can use technologies such as Azure Batch or Azure Machine Learning [11]. Both allow you to create a cluster of identical virtual machines, and have the same code run in parallel on all of them.

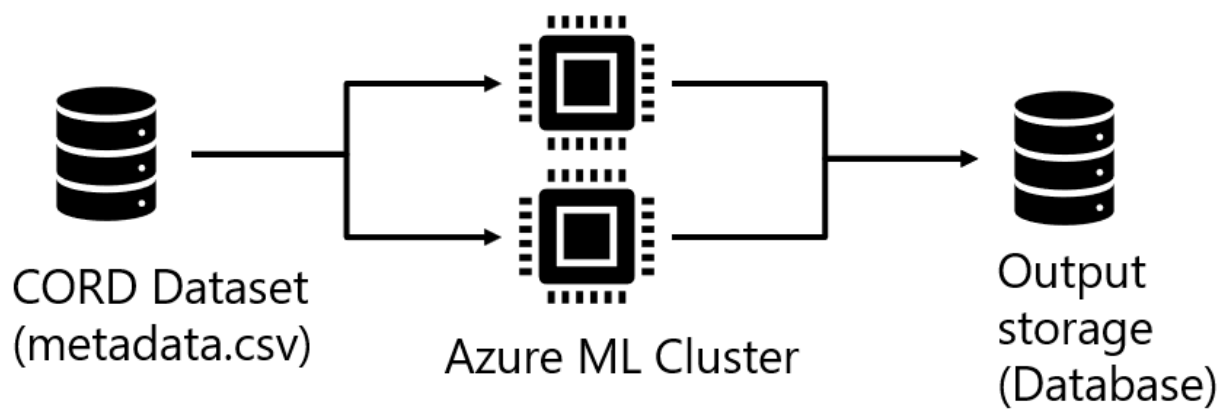

**Fig 3:** Parallel paper processing using Azure Machine Learning Cluster

Azure Machine Learning is a service intended to satisfy all needs of a Data Scientist. It is typically used for training and deploying model and ML Pipelines; however, we can also use it to run our parallel sweep job across a compute cluster. To do that, we need to submit a `sweep_job` experiment. Our experiment is described in more detail in [13].

In our experiment, we created Azure Machine Learning workspace with a cluster of 8 low-performing VMs. We then defined an environment to run the experiment on, based on `mcr.microsoft.com/azureml/base:intelmpi2018.3-ubuntu16.04` container image, and conda configuration files to install required Python SDKs.

We then defined a sweep job to run on the cluster. The job will start `process.py` Python script on each node of the cluster, and pass the number of experiment as well as the dataset and total number of nodes as command-line parameters:

```
experiment_name: sweep_experiment
type: sweep_job
search_space:
  number:
     type: choice
     values: [0,1,2,3,4,5,6,7]
trial:
   command: python process.py
     --number {search_space.number}
     --nodes 8
     --data {inputs.metacord}
   inputs:
      metacord:
        file: azureml:metacord:1
   environment: azureml:cognitive-env:1
   compute:
      target: azureml:AzMLCompute
```

The processing logic will be encoded in the Python script, and will be roughly the following:

```
## process command-line arguments using ArgParse
…
df = pd.read_csv(args.data) # Get metadata.csv into Pandas DF
## Connect to the database
coscli = azure.cosmos.CosmosClient(cosmos_uri, credential=cosmoskey)
cosdb = coscli.get_database_client("CORD")
cospapers = cosdb.get_container_client("Papers")
## Process papers
for i,(id,x) in enumerate(df.iterrows()):
   if i%args.nodes == args.number: # process only portion of record
   # Process the record using REST call (see code above)
   # Store the JSON result in the database
   cospapers.upsert_item(json)
```

For simplicity, we will not show the complete configuration file and script here and refer you to the complete blog post [13], or to the project GitHub repository.

## Using CosmosDB to Store Analytics Result

Using the code above, we have obtained a collection of papers, each having a number of entities and corresponding relations. This structure is inherently hierarchical, and the best way to store and process it would be to use NoSQL approach for data storage. We will use Cosmos DB database to store and query semi-structured data like our JSON collection. The code shown above demonstrates how we can store JSON documents directly into CosmosDB database from our processing scripts running in parallel.

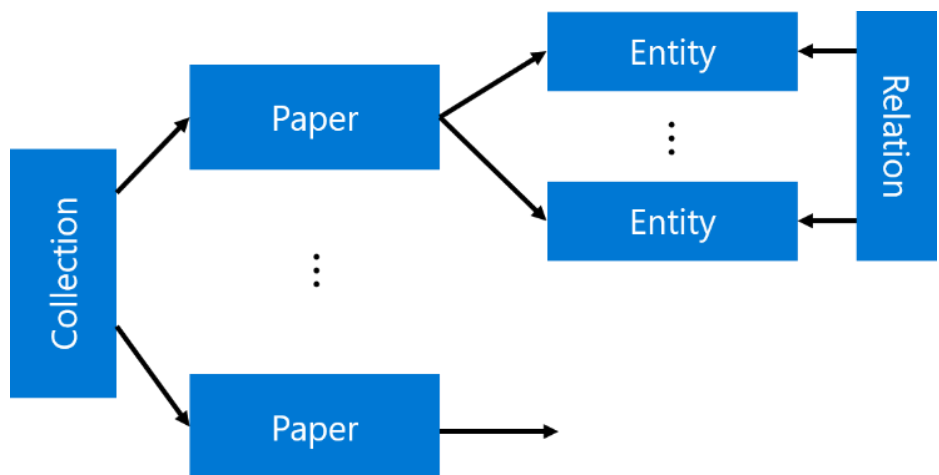

**Fig. 4:** Hierarchical structure of Paper-Entity-Relation relationship.

Now we can use Cosmos DB SQL language in order to query our collection. For example, here is how we can obtain the list of all medications found in the corpus:

```sql
-- unique medication names
SELECT DISTINCT e.text
FROM papers p
JOIN e IN p.entities
WHERE e.category='MedicationName'
```

Using SQL, we can formulate some very specific queries. Suppose a medical specialist wants to find out all proposed dosages of a specific medication (say, **hydroxychloroquine**), and see all papers that mention those dosages. This can be done using the following query:

```sql
-- dosage of specific drug with paper titles
SELECT p.title, r.source.text
FROM papers p JOIN r IN p.relations
WHERE r.relationType='DosageOfMedication'
AND r.target.text LIKE 'hydro%'
```

A more difficult task would be to select all entities together with their corresponding ontology ID. This would be extremely useful, because eventually we want to be able to refer to a specific entity (*hydroxychloroquine*) regardless or the way it was mentioned in the paper (for example, *HCQ* also refers to the same medication). We will use UMLS as our main ontology.

```sql
--- get entities with UMLS IDs
SELECT e.category, e.text,
  ARRAY (SELECT VALUE l.id
         FROM l IN e.links
         WHERE l.dataSource='UMLS')[0] AS umls_id
FROM papers p JOIN e IN p.entities
```

## Creating Interactive Dashboards

While being able to use SQL query to obtain an answer to some specific question, like medication dosages, seems like a very useful tool—it is not convenient for non-IT professionals, who do not have high level of SQL mastery.

| Document.category | Count |
|---|---|
| AdministrativeEvent | 171937 |
| Age | 155693 |
| BodyStructure | 245787 |
| CareEnvironment | 130770 |
| ConditionQualifier | 486762 |
| Date | 42153 |
| Diagnosis | 2477847 |
| Direction | 34398 |
| Dosage | 35550 |
| ExaminationName | 2226245 |
| FamilyRelation | 80454 |
| Frequency | 21274 |
| Gender | 67145 |
| GeneOrProtein | 39782 |
| HealthcareProfession | 120570 |
| MeasurementUnit | 384921 |
| MeasurementValue | 1030936 |
| MedicationClass | 242516 |
| MedicationForm | 7697 |
| MedicationName | 297463 |
| MedicationRoute | 16988 |
| RelationalOperator | 190191 |
| SymptomOrSign | 1117712 |
| Time | 336369 |
| TreatmentName | 1314612 |
| Variant | 4427 |

| Document.text | Count | Document.umls_id |
|---|---|---|
| hydroxychloroquine | 5889 | C0020336 |
| HCQ | 3667 | C0020336 |
| remdesivir | 3188 | C4726677 |
| chloroquine | 2989 | C0008269 |
| Tocilizumab | 2726 | C1609165 |
| lopinavir | 2004 | C0674291 |
| azithromycin | 1980 | C0052796 |
| quarantine | 1890 | C0034386 |
| ritonavir | 1871 | C0292818 |
| cytokine storm | 1499 | |
| insulin | 1202 | C0021641 |
| ribavirin | 1147 | C0035525 |
| dexamethasone | 1118 | C0011777 |
| CQ | 1081 | C0008269 |
| vitamin D | 1081 | C0042866 |
| favipiravir | 1063 | C1138226 |
| heparin | 979 | C0019134 |
| PEDV | 946 | C0318855 |
| Cochrane | 934 | |
| propofol | 879 | C0033487 |
| ozone | 767 | C0030106 |
| Aspirin | 753 | C0004057 |
| cytokine | 750 | C0079189 |
| methylprednisolone | 742 | C0025815 |
| Oseltamivir | 742 | C0874161 |
| COVID-19 | 685 | C5203676 |
| Mpro | 679 | |
| LPS | 674 | C0023810 |
| murine | 667 | C0591833 |
| APA | 632 | |

Entities | Relations | Medicines | +

**Fig. 5:** An interactive entity-relation exploration tool for medical professional based on PowerBI. In this example, you can see all entities of type MedicationName, sorted by number of occurrences, together with their UMLS ID, and you may notice that several entities refer to the same ontology ID (*hydroxychloroquine* and *HCQ*).

To make the collection of metadata accessible to medical professionals, we can use PowerBI tool to create an interactive dashboard for entity/relation exploration.

From this tool, we can make queries similar to the one we have made above in SQL, to determine dosages of a specific medications. To do it, we need to select **DosageOfMedication** relation type in the left table, and then filter the right table by the medication we want. It is also possible to create further drill-down tables to display specific papers that mention selected dosages of medication, making this tool a useful research instrument for medical scientist.

# Getting Visual Insights

The most interesting part of the story, however, is to automatically draw some visual insights from the text, such as the change in medical treatment strategy over time. To do this, we need to write some more code in Python to do proper data analysis and visualization. The most convenient way to do that is to use Notebooks embedded into Cosmos DB:

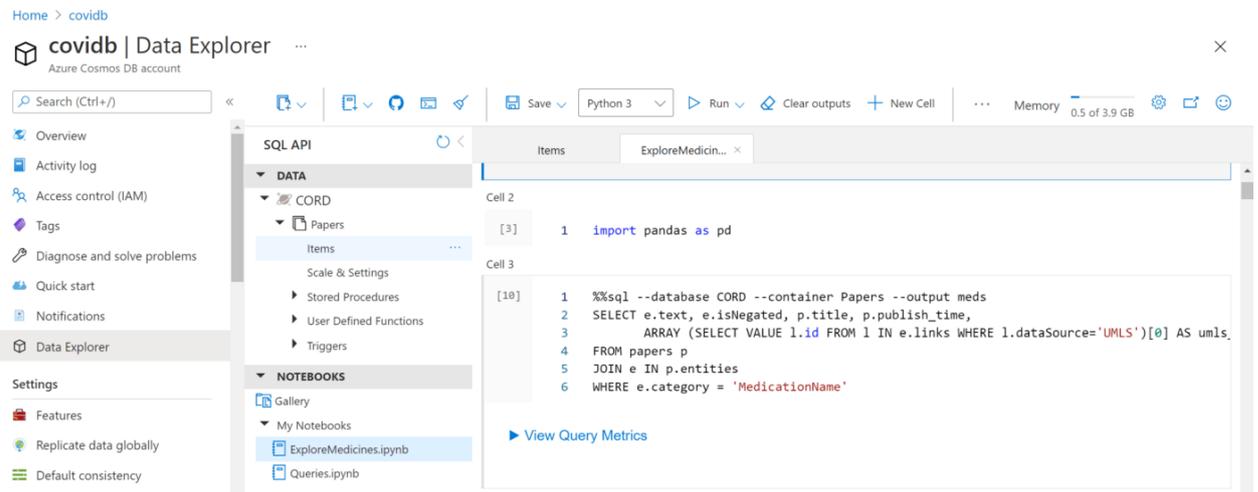

**Fig. 6:** Jupyter Notebooks Embedded into Cosmos DB Data Explorer.

Those notebooks support embedded SQL queries; thus, we are able to execute SQL query, and then get the results into Pandas DataFrame, which is Python-native way to explore data:

```
%%sql --database CORD --container Papers --output meds
SELECT e.text, e.isNegated, p.title, p.publish_time,
       ARRAY (SELECT VALUE l.id FROM l
              IN e.links
              WHERE l.dataSource='UMLS')[0] AS umls_id
FROM papers p
JOIN e IN p.entities
WHERE e.category = 'MedicationName'
```

Here we end up with `meds` DataFrame, containing names of medicines, together with corresponding paper titles and publishing date. We can further group by ontology ID to get frequencies of mentions for different medications:

```
unimeds = meds.groupby('umls_id') \
            .agg({'text' : lambda x : ','.join(x),
                  'title' : 'count',
                  'isNegated' : 'sum'})
unimeds['negativity'] = unimeds['isNegated'] / unimeds['title']
unimeds['name'] = unimeds['text'] \
            .apply(lambda x: x if ',' not in x
                             else x[:x.find(',')])
unimeds.sort_values('title',ascending=False).drop('text',axis=1)
```

This gives us the following table:

| umls_id | title | isNegated | negativity | name |
|---|---|---|---|---|
| C0020336 | 4846 | 191 | 0.039414 | hydroxychloroquine |
| C0008269 | 1870 | 38 | 0.020321 | chloroquine |
| C1609165 | 1793 | 94 | 0.052426 | Tocilizumab |
| C4726677 | 1625 | 24 | 0.014769 | remdesivir |
| C0052796 | 1201 | 84 | 0.069942 | azithromycin |
| ... | ... | ... | ... | ... |
| C0067874 | 1 | 0 | 0.000000 | 1-butanethiol |

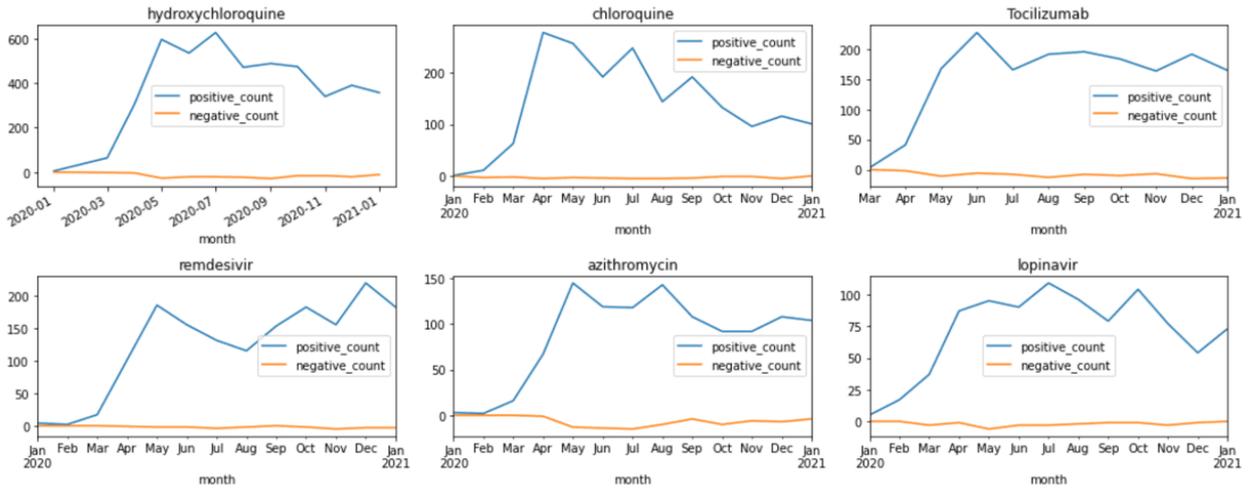

**Fig. 7:** Visualization of change in treatment strategies by observing the number of occurrences. Blue line shows occurrences in positive context, while yellow line shows negative occurrences (lower values correspond to more negative occurrences).

Further transforming the data, we can get the number of monthly mentions for top medications, to see how it changes over time. The result is presented in Fig. 7. From those graphs you can see that hydroxychloroquine was the most discussed medicine in the beginning of the pandemic, and then number of mentions dropped. On the other hand, medicines such as remdesivir show stable rise in the number of mentions.

While absolute numbers of mentions give us some idea on the current directions of research, it would be more interesting to visualize the relative number of mentions of medicines. In Fig 8 we take top 12 most frequently mentioned medications, and visualize their relative frequency of mentions compared to other medications from this list. We can effectively see how percentages of mentions of different medications change over time.

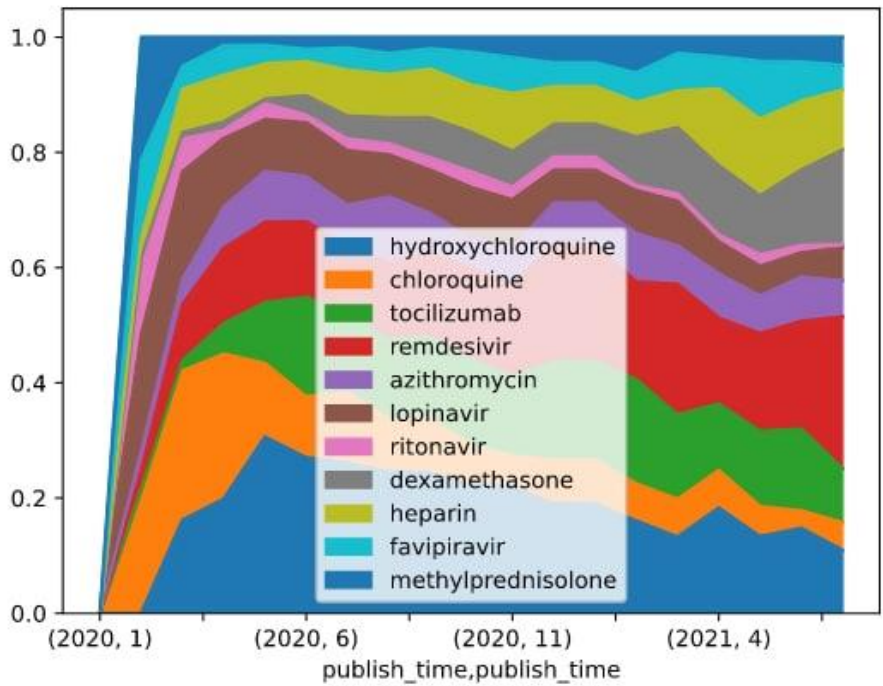

**Fig. 8:** Visualization of change in treatment strategies in percentage of total number of mentions for top 12 medications.

## Visualizing Terms Co-Occurrence

Another interesting insight is to observe which terms occur frequently together. To visualize such dependencies, there are two types of diagrams:

- **Sankey diagram** allows us to investigate relations between two types of terms, eg. diagnosis and treatment
- **Chord diagram** helps to visualize co-occurrence of terms of the same type (eg. which medications are mentioned together)

To plot both diagrams, we need to compute **co-occurrence matrix**, which in the row `i` and column `j` contains number of co-occurrences of terms `i` and `j` in the same abstract (one can notice that this matrix is symmetric). For the visualization to be more clear, we select relatively small number of terms from our ontology, grouping some terms together if needed. To plot the Sankey diagram, we use Plotly graphics library.

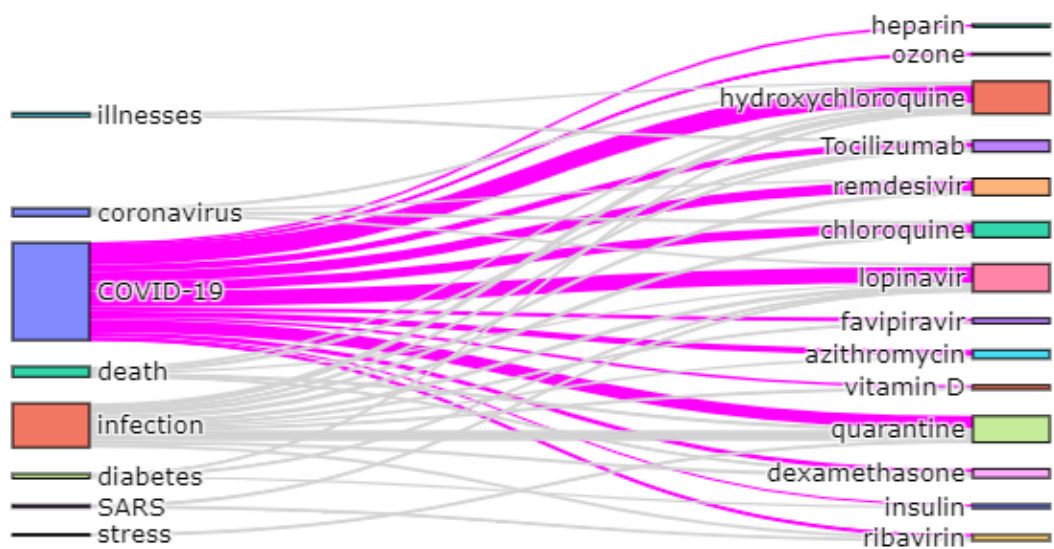

**Fig.9:** Diagram showing the frequency of co-occurrences of different diagnoses (on the left) and medications (on the right).

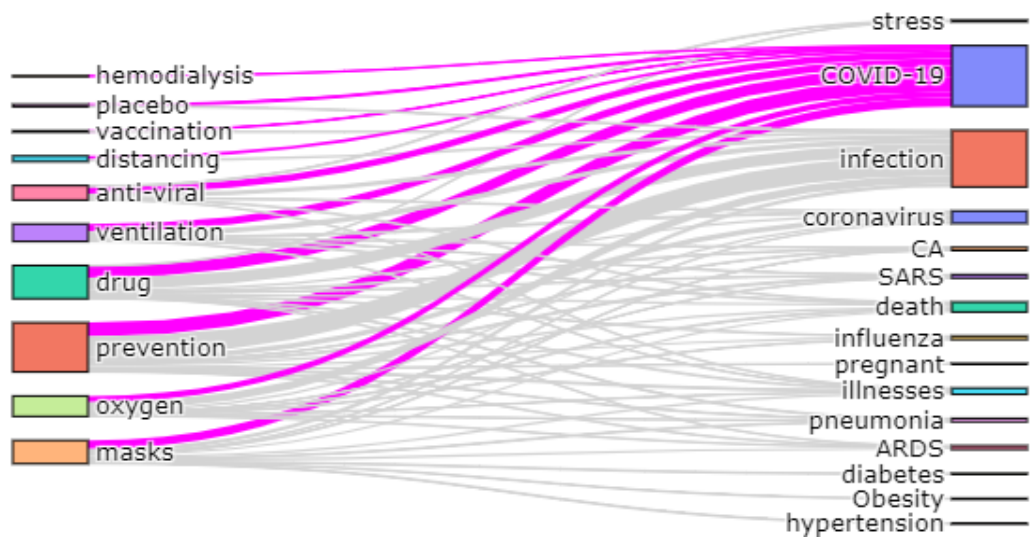

**Fig.10:** Diagram showing the frequency of co-occurrences of different treatments (on the left) and diagnoses (on the right).

For visualizing co-occurrences of entities of the same type – eg. different medications, we can plot a chord diagram. We use a libary called Chord, and the same function as above to populate co-occurrence matrix, passing the same ontology twice.

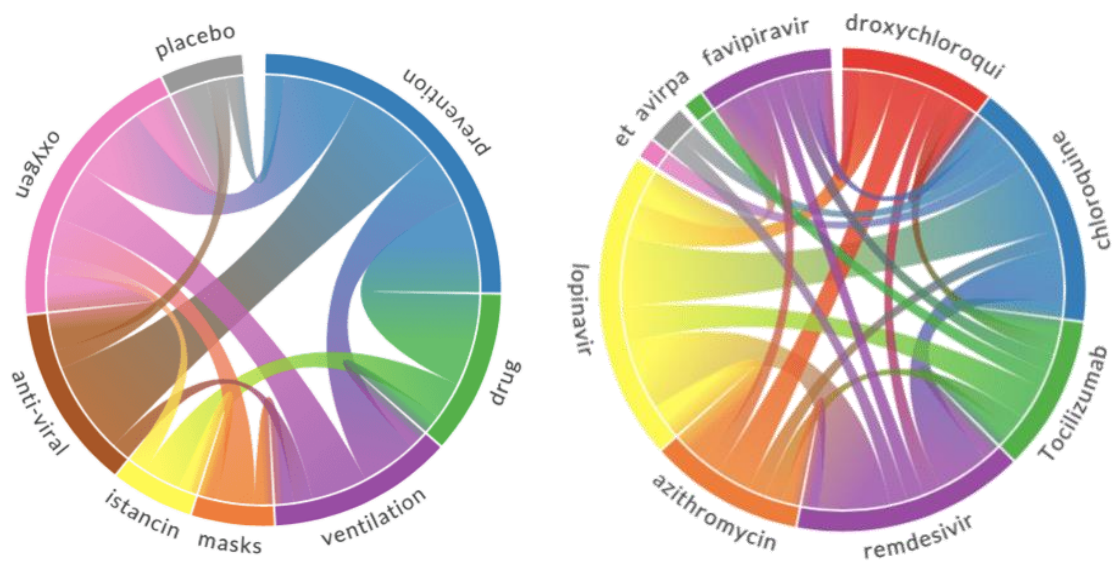

**Fig 11:** Co-occurrence chord diagrams for treatments (on the left) and medications (on the right).

Please note that on the right diagram of co-occurrences of medications we can clearly see some well-known combinations of medications, such as **hydroxychloroquine + azitromycin**, which were included into standard treatment strategy [14]. We can also see that chloroquine and lopinavir are frequently mentioned together, but that does not necessarily mean that they are used together (for the counter-example, see [15]). This demonstrates that we need to perform deeper text analysis to understand the nature of co-occurrence of different terms in the abstract.

## Conclusion

In this paper, we have described the architecture of a proof-of-concept system for knowledge extraction from large corpora of medical texts. We use Text Analytics for Health to perform the main task of extracting entities and relations from text, and then a number of cloud services together to build a query tool for medical scientist and to extract some visual insights.

For further research, it would be interesting to switch to processing full-text articles in addition to abstracts, in which case we need to think about slightly different criteria for co-occurrence of terms (eg. in the same paragraph vs. the same paper).

The same approach can be applied in other scientific areas, but we would need to be prepared to train a custom neural network model to perform entity extraction (and for that we might need to both fine-tune BERT model on text from the problem domain, and train a model based on fine-tuned BERT feature extractor to perform NER (for that we would need relatively large dataset of labeled entities).

We hope that the proposed approach would be nevertheless transferable to different problem domains, and that the codebase that we provide can be used as a starting point for further research in the area of using natural language processing machinery to gain insights from large text corpora.